\documentclass[letterpaper, 10 pt, conference]{ieeeconf}
\IEEEoverridecommandlockouts  
\usepackage[utf8]{inputenc}
\usepackage{graphicx,epstopdf}
\usepackage{multicol,color}
\newcommand{\RNum}[1]{\uppercase\expandafter{\romannumeral #1\relax}}
\usepackage{slashbox}
\usepackage[ruled,vlined]{algorithm2e}
\title{\LARGE \bf OFFSEG: A Semantic Segmentation Framework For Off-Road Driving}
\author{Kasi Viswanath{$^*$}, Kartikeya Singh{$^*$}, Peng Jiang, P.B. Sujit and Srikanth Saripalli%
\thanks{Kasi Viswanath, Kartikeya Singh and P.B. Sujit are with the Department of Electrical Engineering and Computer Science, IISER Bhopal, Bhopal - India. e-mail:(kasi18,kartikeyas,sujit)@iiserb.ac.in}
\thanks{Peng Jiang and Srikanth Saripalli are with the Department of Mechanical Engineering, Texas A\&M University, College Station, Texas, TX--  77843-3123. e-mail:{(maskjp,ssaripalli)@tamu.edu}}
\thanks{{$^*$} Kasi Viswanath and Kartikeya Singh are equal contributors.}}
\begin{document}
\maketitle
\thispagestyle{empty}
\pagestyle{empty}
\begin{abstract}
Off-road image semantic segmentation is challenging due to the presence of 
uneven terrains, unstructured class boundaries, irregular features and strong textures. These aspects  affect the perception of the vehicle from which the information is used for path planning. 
Current off-road datasets exhibit difficulties like class imbalance and understanding of varying environmental topography.  
To overcome these issues we propose a framework for off-road semantic segmentation called as OFFSEG that involves (i) a pooled class semantic segmentation with  four classes (sky, traversable region, non-traversable region and obstacle) using state-of-the-art deep learning architectures (ii) a colour segmentation methodology to segment out specific sub-classes (grass, puddle, dirt, gravel, etc.) from the traversable region for better scene understanding.  
The evaluation of the framework is carried out on two off-road driving datasets, namely, RELLIS-3D and RUGD. We have also tested proposed framework in IISERB campus frames. The  results show that OFFSEG achieves good performance and also provides detailed information on the traversable region. 
\end{abstract}
\section{Introduction}


Autonomous off-road driving has wide range of applications like inspection, exploration, rescue, reconnaissance missions, etc. Off-road environments are often texture rich with indefinite boundaries and less detailed than urban environment.  
Non-uniform terrain description makes off-road environment more difficult to understand from perception perspective for a robust autonomous driving system.

\begin{figure}
\centering
    \includegraphics[width=8cm, height=7cm]{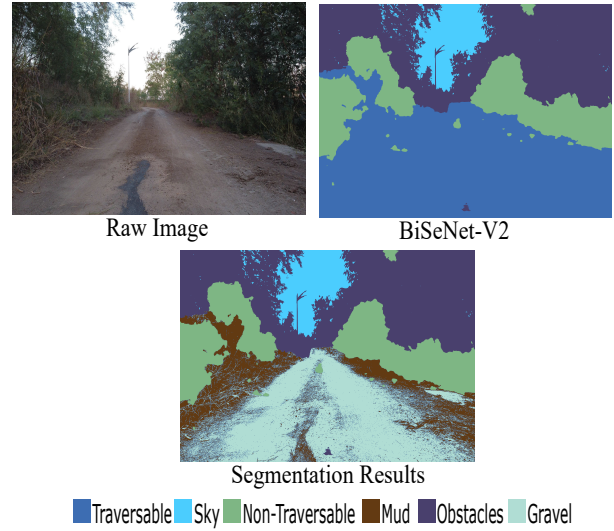}
    \caption{Segmentation results on IISERB campus frames (top left) Raw image taken from the vehicle (top right) Semantic segmentation of  four clases using BiSeNetV2 (below) Color segmentation on the traversable region providing additional information about the traversable region  }
     \label{fig:iiserb1}
\end{figure}

On road driving has received significant attention in the domain of autonomous driving in terms of datasets for segmentation. The datasets \cite{yu2020bdd100k}\cite{apolloscape_arXiv_2018}\cite{cordts2016cityscapes} are available for the semantic scene understanding of the on-road environment as well as the state-of-the-art benchmarks for these environments. The on-road driving is mainly for urban environments. Compared to urban environments, off-road environments have unstructured class boundaries, uneven terrain, strong textures, and irregular features that hinders the direct transfer of models between the environments. Moreover, there are large differences in class distributions across distinct off-road environments. Thus, there is a need to develop a framework that provides feature rich semantic information to the vehicle for making better decisions while driving in off-road environments. For example, consider Figure \ref{fig:iiserb1} that provides a detailed information on the traversable region like mud and gravel which could be used by the path planning module of an autonomous vehicle for better planning as compared to the traversable region without any additional information.

The limited availability of off-road environment based datasets is another challenge which hinders the progression of off-road autonomous driving domain. For the robotic navigation in the off-road environment there are three main datasets (1) RELLIS--3D \cite{jiang2020rellis3d}, (2) RUGD \cite{8968283}, and (3) DeepScene \cite{valada16iser}. RELLIS--3D \cite{jiang2020rellis3d} is a multimodal dataset collected in an off-road environment, which contains annotations for 13,556 LiDAR scans and 6,235 images where ground truth in terms of annotated labels are provided. 
RUGD \cite{8968283} dataset gives a rich ontology and large set of ground truths of 7546 annotations with 24 classes. 
However, in both RELLIS-3D and RUGD, classes like log, pole, water exhibit low pixel density resulting in class imbalance and hence low mIoU. 


An interesting aspect of off-road driving is that unlike on-road scenes, which has detailed classes like signboard, traffic lights, etc,  off-road environments require less features which allows us to pool the classes in RELLIS-3D and RUGD dataset to four classes namely traversable, non-traversable, obstacles and sky. The motive behind clustering classes into four was to group classes according to their semantic contributions in the environment. The sky class explicitly includes the region present in sky. The traversable class includes all possible traversable regions present in the datasets. The non-traversable class includes all surface regions which are not traversable and do not act as an obstacle during the off-road navigation. Lastly, the obstacle class explicitly includes all the possible obstacles present in the dataset. 
\begin{figure*}
    \centering
    \includegraphics[width=18cm, height=4cm]{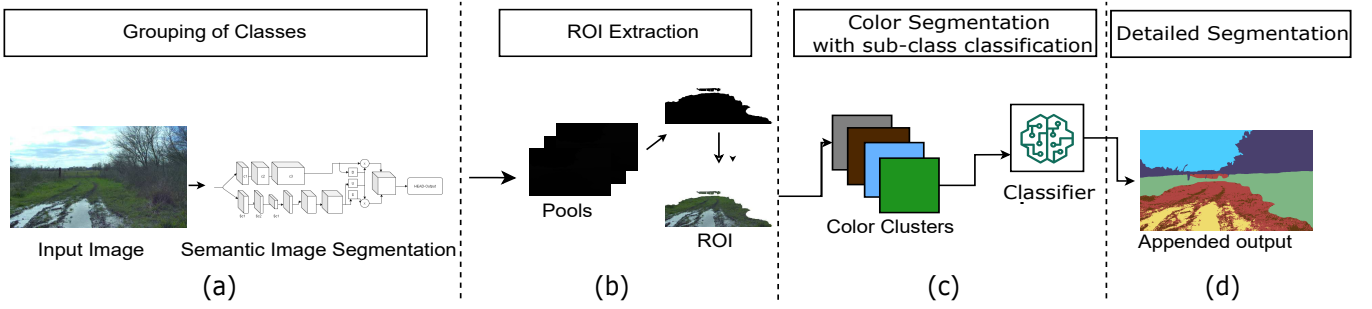}\\
    \caption{OFFSEG consists of two stages. First, pooling of different classes into four and performing semantic image segmentation. Second, the RoI (region of interest) obtained from the segmentation passes through the color segmentation algorithm which segments and classifies sub classes like grass, mud, puddle, etc and append them as a final output.}
     \label{fig:framework}
\end{figure*}
By grouping the classes into four, we resolve the class imbalance issue adequately. The traversable class can include additional information like, dirt, mud, gravel, etc. These additional details can play an important role in determining the drivable path during autonomous off-road driving.   


A transfer learning framework with semantic segmentation  for off-road environments was developed  in \cite{sharma2019semantic}. However, the approach has lower performance when providing  finer features like grass, gravel, bush, etc., which are essential attributes to perform robust robotic navigation in  off-road environments. Nefian and Bradski \cite{4107207} use a hierarchical  Bayesian network approach to detect driving regions. However, the approach does not provide detailed features and the environment is not heterogeneous. 

The main contributions of this paper are 
\begin{itemize}
    \item Development of a novel, simple and efficient semantic segmentation framework for off-road driving OFFSEG. 
    \item We exploit the off-road driving requirement by pooling the 20 classes in RELLIS-3D and 24 classes in RUGD into 4 classes and mitigate the class imbalance issue.
    \item We use color layers \cite{lee2021new} and K-Means \cite{ding2015yinyang} clustering to determine RGB clusters \cite{khairudin2021fast} \cite{naji2012skin} to find finer details on the traversable region.
    \item We compare the results of BiSeNetV2 and HRNETV2+OCR on RELLIS-3D and RUGD. The mIoU of OFFSEG cannot be compared with the results of HRNETV2+OCR and GSCNN from RELLIS-3D because of the 20 pooled to 4 in our proposed framework.
     Similarly for RUGD the 24 classes were pooled to 4 hence we cannot make a direct inference of the results with benchmarks. 
    \item We also test OFFSEG on IISERB campus frames that we recorded from IISERB campus. The outputs obtained from all the datasets shows detailed segmentation of all classes.
    \item We test OFFSEG  on NVIDIA Jetson AGX Xavier \cite{jetsonAGX} to record the inference speed of our framework.
\end{itemize}
The rest of the paper is organized as further. The Section \RNum{2} describes the OFFSEG framework developed in this paper. The results obtained from the framework are given in Section \RNum{3}. We provide conclusion and future work in Section \RNum{4}.


\section{Methodologies}
OFFSEG consists of two stages: semantic segmentation for four classes and colour segmentation of traversable region. The input is a raw RGB image and the output is a pixel-wise annotated RGB image. An overview of OFFSEG is represented in Figure \ref{fig:framework}.
\subsection{Semantic segmentation}
Semantic Segmentation is a method of labeling the class of each pixel in an image. Traditionally, image segmentation was carried out by threshold selection, region growing etc. Recent developments of the Convolutional Neural Network
(CNN)\cite{garciagarcia2017review} yielded in faster and accurate state-of-the-art segmentation architectures. Most architectures consist of an encoder-decoder structure which downsamples the input to extract features and then upsamples with a pooling layer. This results in loss of spatial details. Architectures like HRNET\cite{WangSCJDZLMTWLX19} adopts a high resolution multiple branch to recover the spatial information. 

As shown in Figure \ref{fig:framework}.a, the first stage of our framework is to perform semantic segmentation on 4 classes. The 20 classes in RELLIS-3D and 24 classes in RUGD datasets were re-categorised into four classes, 1) sky, 2) traversable, 3) non-traversable 4) obstacle.

From RELLIS-3D 6 classes were pooled to traversable, 3 were pooled to non-traversable and 10 were pooled to Obstacles. As the RELLIS-3D dataset had $94\% $ of the pixels distributed between sky, grass, tree and bushes, pooling them into four different classes solved the problem of class imbalance issue as shown in Figure \ref{fig:log} . The pixel-wise annotation of the classes from RELLIS-3D and RUGD were then converted into these four classes for training on the semantic segmentation network.
\subsubsection{Sub class featuring of traversable region}
The traversable class includes sub-classes like puddle, mud, dirt, gravel. A proper sub-class distribution is shown in Table \RNum{1} . These sub-classes play an important role in determining path during the robotic navigation in the off-road environment. Another reason to consider only the traversable class as our RoI is to ignore all other unusable sub-classes present in the environment which are not necessary for determining traversable path in autonomous driving like pole, bush, etc. These instances do not require fine segmentation to achieve. We obtain detailed segmentation of these sub-classes in Section \RNum{3}.

\subsection{Color segmentation and sub-class Classification}
\begin{figure}
    \centering
    \includegraphics[width=8cm, height=6cm]{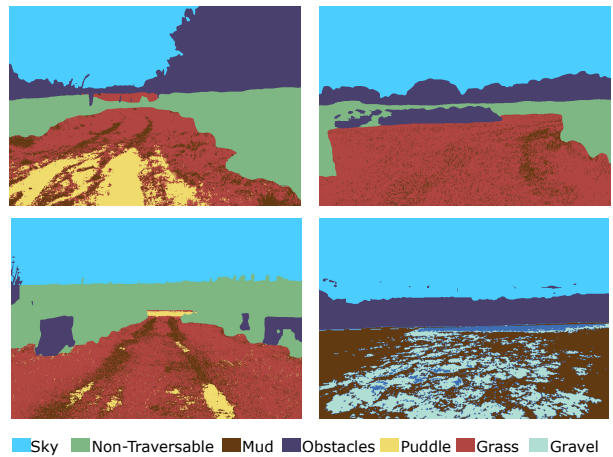}
    \caption{Final segmentation of classes (sky, traversable, non-traversable, obstacles) and sub-classes (mud, puddle, grass, gravel) in a,e,b,d frames respectively.}
    \label{fig:final}
\end{figure}

K-Means algorithm has been used to extract the color pools from the output obtained in previous section. Color pools are used to distinguish between several components present in an off-road environment. Each cluster obtained from the centroid has been transferred into the classification model which gives us the mapping of the required sub-class in our region of interest as shown in Figure \ref{fig:framework}.b. The color segmentation algorithm extracts the color masks from the image and inputs these masks into our classification model. The classifier classifies the sub-classes in terms of color clusters and determines our sub-classes like mud, puddle, grass, water, etc as shown in Figure \ref{fig:framework}.c. Next, these obtained masks are appended on our segmentation result which was obtained from our semantic segmentation resulting in final segmentation as shown in Figure \ref{fig:final}.
\begin{center}
\begin{table}
\centering
\begin{tabular}{ |p{1.4cm}||p{1.9cm}|p{1.9cm}|p{1.8cm}|  }
 \hline
 \multicolumn{4}{|c|}{\textbf{Class Distribution for $\textbf{RELLIS-3D}$ and $\textbf{RUGD}$ }} \\
 \hline
 \textbf{Sky}&\textbf{Traversable}&\textbf{Non Traversable}&\textbf{Obstacles}\\
 \hline
 Sky[RE,RU]&Grass[RE,RU]&Bush[RE,RU]&Vehicle[RE,RU]\\
 &Dirt[RE,RU]&Void[RE]&Barrier[RE]\\
 &Asphalt[RE,RU]&Water[RE,RU]&Log[RE,RU]\\
 &Concrete[RE,RU]&Deep Water[RE]&Pole[RE]\\
 &Puddle[RE]&&Object[RE]\\
 &mud[RE]&&Building[RE,RU]\\
 &Sand[RU] &&Person[RE,RU]\\
 &Gravel[RU]&&Fence[RE,RU]\\
 &Mulch[RU] &&Tree[RE,RU]\\
 &Bridge[RU]&&Rubble[RE]\\
 &Rockbed[RU]    &     &Pole[RU]     \\
 &     &    & Container[RU]     \\
 &     &     &Bicycle[RU]     \\
 &    &     &Sign[RU]     \\
 &     &     &Rock[RU]     \\
 &     &     &Table[RU]     \\
 \hline
\end{tabular}

\caption{class distribution of the polled classes from RELLIS-3D[RE] and RUGD[RU]}\vspace{-1cm}
\label{table:3}
\end{table}
\end{center}
\subsubsection{Data pre-processing and data generation for classification}
The traversable class provides refining of the region of interest up to an extent and provides us the area space to extract the training candidates from the region of interest. The training samples in RELLIS-3D came out as 6 classes- grass, mud, puddle, dirt, asphalt, concrete. and in RUGD as- dirt, sand, grass, water, asphalt, gravel, mulch and concrete. We have prepared an image oriented dataset for the training of a classification model. This dataset comprises of the detailed sub classes present in our traversable region.

\subsubsection{Training of classification model}
Outputs from colour segmentation needs to be classified into different sub-classes in the traversable region. The classifier differentiates the masks extracted from K-Means clustering and assigns the respective classes to them. Table \RNum{1} shows the different sub-classes present in the traversable region of both RELLIS-3D and RUGD dataset. 
 
\section{Results}
In this Section, we have experimented OFFSEG on two state-of-the-art off-road datasets RELLIS-3D and RUGD. We have also tested OFFSEG on IISERB campus frames.

\subsection{Segmentation}
The evaluation of image semantic segmentation of the converted classes of RELLIS-3D and RUGD were done using two state-of-the-art architectures: BiSeNetV2\cite{yu2020bisenet} and HRNETV2\cite{WangSCJDZLMTWLX19}+OCR\cite{yuan2020objectcontextual}. 
BiSeNetV2 consists of two branches: detail branch and semantic branch. The detail branch extracts spatial details consisting of low-level information and uses shallow layers with wide channels. Meanwhile semantic branch extracts high-level semantics employing low channel capacity. Then an aggregation layer merges extracted features from the two branches and upsample the output from aggregation layer.

HRNETV2+OCR consists of a High-Resolution Network which acts as a backbone and Object-Contextual Representations (OCR) to enhanced pixel representation of objects. Unlike other segmentation models HRNET maintains high resolution throughout the model avoiding the downsample and upsample process. OCR aggregated the features extracted from HRNET to improve pixel representation.
We used 3,302 images for training set, 983 images for validation set and the testing with 1672 images for the RELLIS-3D. For RUGD, we used 4732 images for training set, 932 images for validation set and 1827 images for testing set.
\begin{figure}
    \includegraphics[width=8cm, height=6cm]{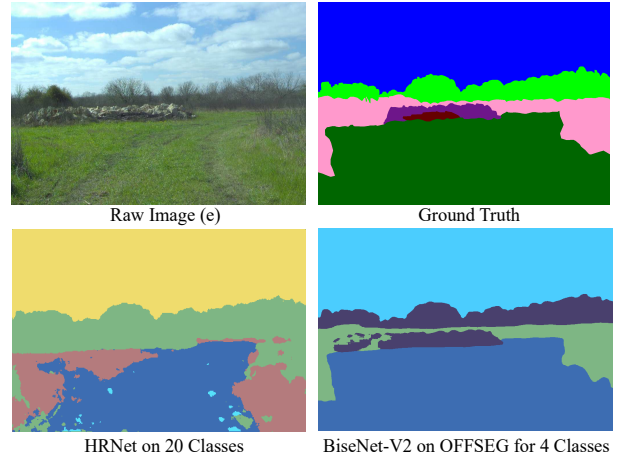}
    \caption{Segmentation results from OFFSEG for four class model have been compared with the HRNET 20 class model}
     \label{fig:log}
\end{figure}
\subsubsection{Quantitative analysis of the architectures used for segmentation}
The results obtained from our OFFSEG can be seen in Figure \ref{fig:segmentation}. The individual IoU breakout of four classes is given in Table \RNum{2}. The mean IoU \cite{everingham2015pascal} for the datasets has given by:- 
\begin{equation}
    mIoU = \frac{1}{n} \sum_{n=1}^{n} Z
\end{equation}
\begin{equation}
\resizebox{1\hsize}{!}{$Z=\frac{(TruePositive)_n}{(TruePositive)_n+(FalsePositive)_n+(FalseNegative)_n}$}
\end{equation}
where $n$ is the number of classes. 

From Table \RNum{2}, the mean IoU obtained for RELLIS-3D on BiSeNetV2 and HRNETV2+OCR were 86.61\% and 80.82\% respectively. The mean IoU obtained for RUGD on BiSeNetV2 and HRNETV2+OCR were 80.17\% and 84.49\% respectively.

The results obtained in Figure \ref{fig:log} shows the prediction of BiSeNetv2 on a RELLIS-3D frame which contains the class log. The obstacle class in the prediction covers the most of the log ground truth labels inferring higher predictions than the prediction of HRNETV2+OCR trained on 20 classes for log which had 0.0\% IoU.
\subsection{Clustering}
We obtain color clusters using K-Means algorithm. The set of random k-points has been assigned with the closest centroid from the image which further combines these centroids into separate clusters. By adopting an iterative approach, we obtain a set of all possible color clusters present in the RGB layers. The number of clusters to be obtained from the image has been set manually, but depends upon the versatility of the color points present in the image which could be used to classify the sub-classes from the region of interest.

\subsection{Color segmentation and sub-class Classification}
\begin{figure}
\centering
    \includegraphics[width=8cm, height=6cm]{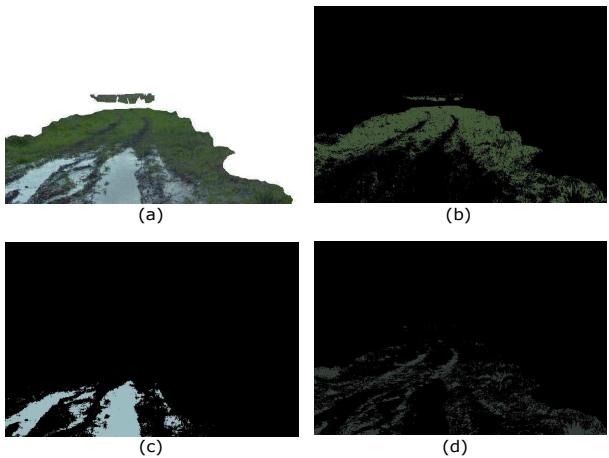}
    \caption{Color segmentation results on RELLIS-3D (a) Traversable class obtained as RoI from segmentation (b) grass, (c) puddle, (d) mud obtained from color segmentation from RoI.  }
    \label{fig:color2}
\end{figure}

\begin{figure}
    \centering
    \includegraphics[width=8cm, height=6cm]{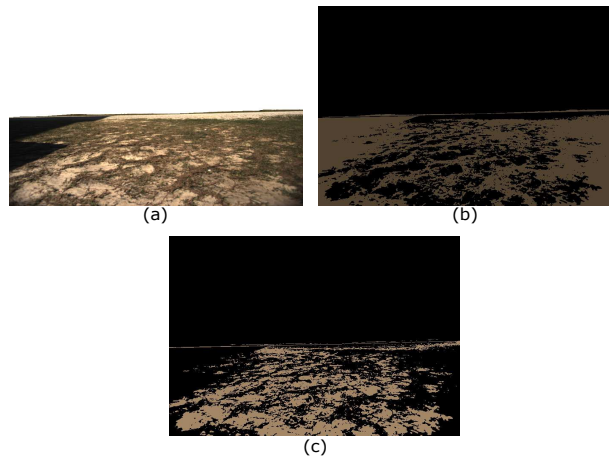}
    \caption{Color segmentation results on RUGD (a) Traversable class obtained as RoI from segmentation (b) mulch, (c) gravel obtained from color segmentation from RoI.}
    \label{fig:color1}
\end{figure}

The color masks obtained after applying color segmentation on our RoI are shown in Figure \ref{fig:color2} and Figure \ref{fig:color1}. The classification between fine details of traversable region shows the accuracy of OFFSEG.
We trained a classification model using transfer learning with MoblieNetV2\cite{sandler2019mobilenetv2} as the classifier model pretrained on ImageNet dataset\cite{imagenet_cvpr09}. The training inputs are classes listed in the traversable region of Table \RNum{1} from both RELLIS-3D and RUGD dataset which play crucial role in detailing of the segmented portion. The classifier was trained on 23,967 images for 9 classes which achieved a mean accuracy of 97.3\% and the outputs obtained from the model transfer knowledge into our color segmentation algorithm which appends only classified sub-classes into our final result. 

Note that, quantitative analysis for the colour segmentation would lead to inaccurate results as the ground truth for classes in traversable region is very vague whereas the outputs in our approach are more feature rich with distinct boundaries.
The detailed outputs obtained expands the application space of the model.
\subsection{Inference speed}
\begin{figure}
    \centering
    \includegraphics[width=8.5cm, height=7cm]{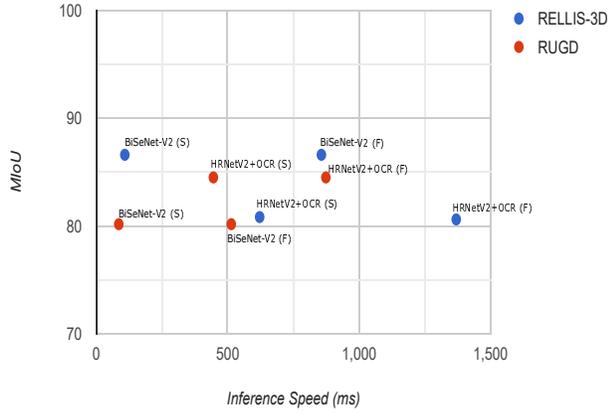}
    \caption{OFFSEG Performance: Segmentation(S) and OFFSEG(F). The testing was performed on Jetson AGX Xavier platform. }
     \label{fig:inf}\vspace{-2mm}
\end{figure}
The inference speed of the whole framework with BiSeNetV2 and HRNETV2+OCR as segmentation model was tested on Jetson AGX Xavier platform \cite{jetsonAGX}. The input resolution of RELLIS-3D is 1024*640 and RUGD is 688*550. The performance graph in Figure \ref{fig:inf} represents the mIoU obtained within corresponding inference speed. This distinguish between the performance of the two architectures used in the framework for testing.

\begin{figure*}
    \centering
    \includegraphics[width=17cm, height=11cm]{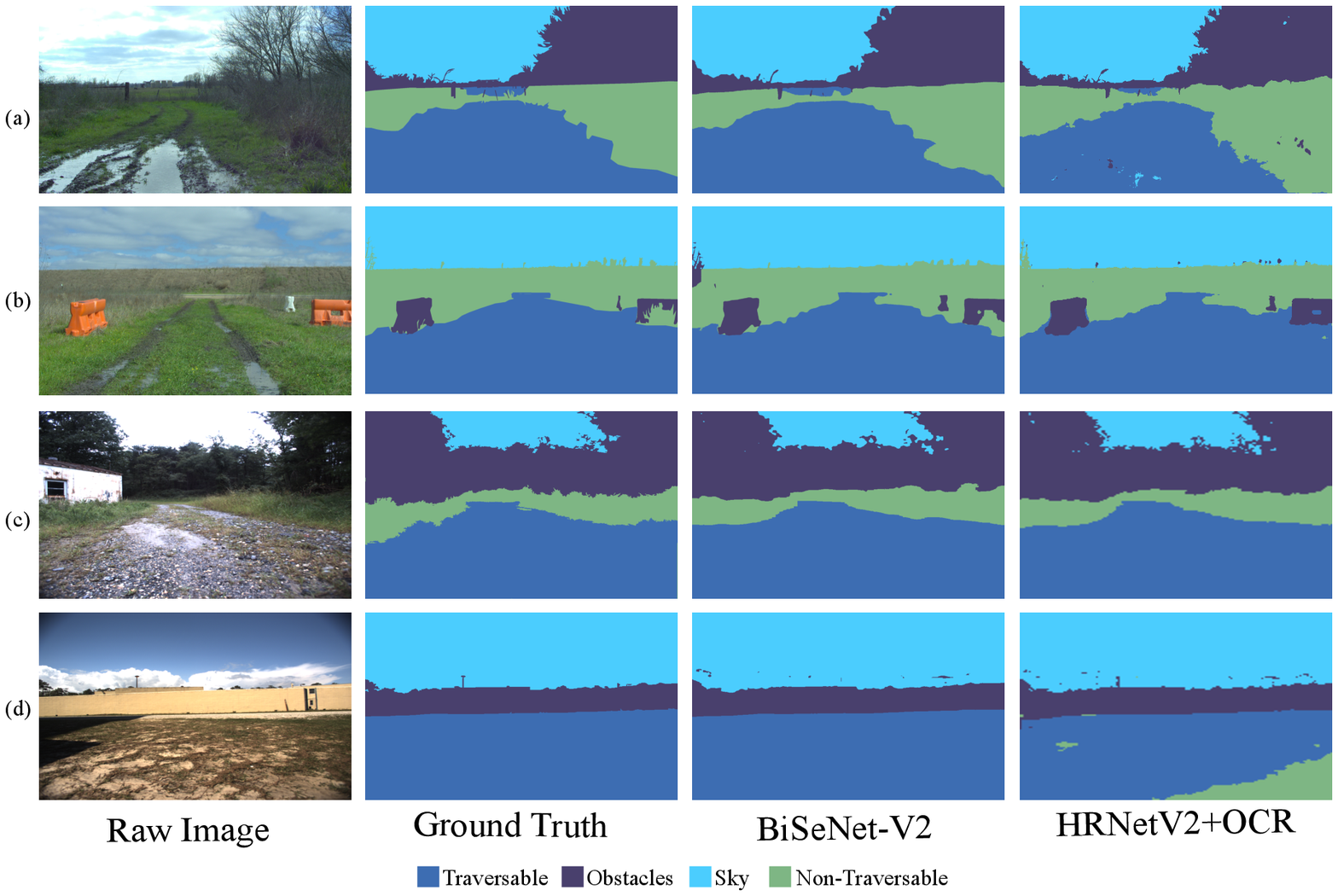}
    \caption{Segmentation results from BiseNetV2 and HRNETV2+OCR}
    \label{fig:segmentation}
    \vspace{4.00mm}
    
    \includegraphics[width=17cm, height=7.3cm]{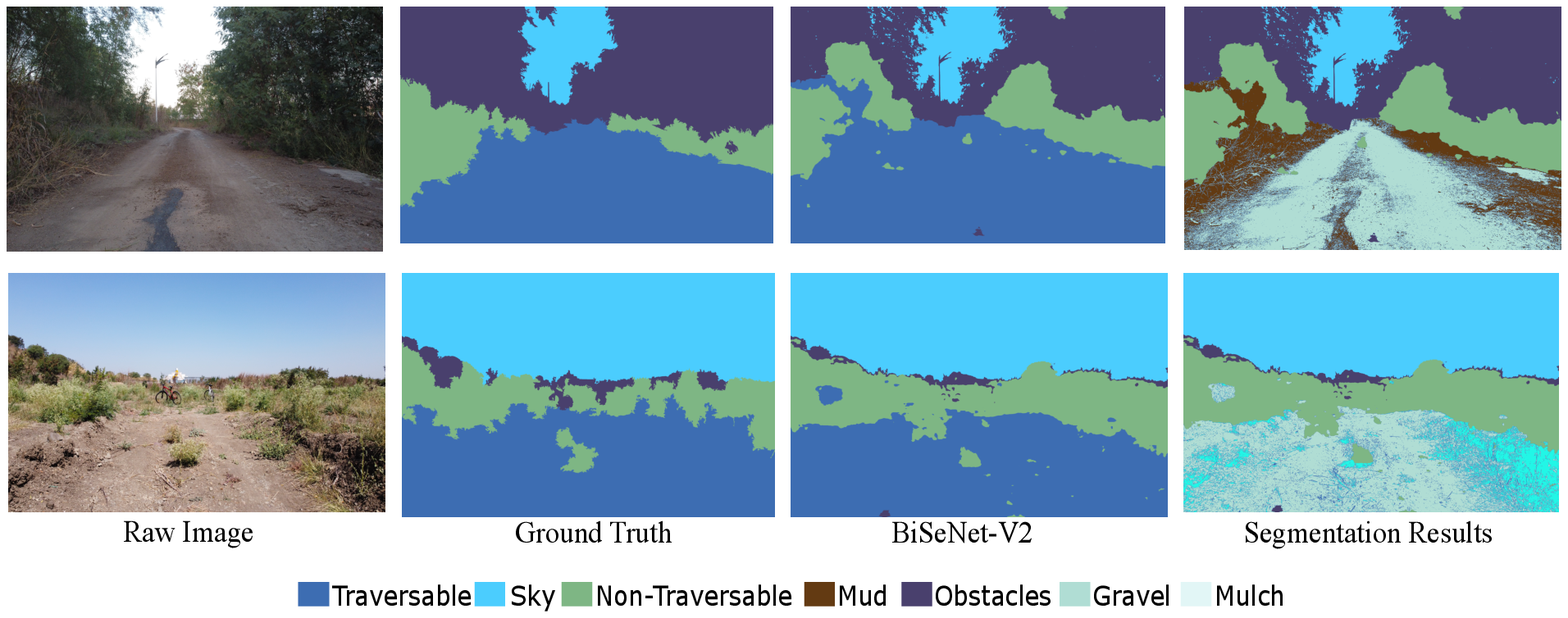}
    \caption{Segmentation results on IISERB campus frames}
     \label{fig:iiserb}
\end{figure*}

\begin{table}
\centering
\resizebox{\columnwidth}{!}{%
\begin{tabular}{|l|c|c|c|c|c|}\hline
\backslashbox{Models}{Classes}
&{Sky}&{Traversable}&{Non-Traversable}&{Obstacles}&{\textbf{mIoU}}\\\hline
BiSeNet-V2 [RELLIS-3D] &97.09\%&92.30\%&77.12\%&79.93\%&86.61\%\\\hline
HRNETV2 [RELLIS-3D] &96.85\%&86.04\%&66.22\%&74.18\%&80.82\%\\\hline
BiSeNet-V2 [RUGD] &90.85\%&91.83\%&47.81\%&90.20\%&80.17\%\\\hline
HRNETV2 [RUGD] &92.27\%&94.18\%&59.92\%&91.60\%&84.49\%\\\hline
BiSeNet-V2 [IISERB] &95.71\%&85.93\%&49.31\%&66.58\%&74.38\%\\\hline
\end{tabular}%
}
\label{table:3}
\caption{mIoU of experiment results}\vspace{-1cm}
\end{table}

\subsection{Testing of framework on IISERB campus frames}
We further evaluated OFFSEG in an untrained environment data from IISERB campus. The Figure \ref{fig:iiserb} shows the  raw images  from IISERB campus. The frames used for testing includes combination of sub-classes present in RELLIS-3D and RUGD datasets. The frames were recorded in a sequential manner using Dji Mavic Mini\cite{dji} from the altitude of 1.8 meters. The ground truth is generated using LabelBox\cite{labelbox} for the evaluation. The BiSeNetV2 model trained using RUGD dataset was used for testing and obtained an mIoU of 74.38\% and the individual class IoU breakout is shown in Table \RNum{3}. The outputs obtained from the model followed by our color segmentation algorithm were adequate to understand the detailed sub-classes (i.e gravel, mulch, mud, etc). From the outputs, we observe that the color segmentation tends to be a very effective mechanism to classify among different sub classes (mud, mulch, gravel) present in the traversable region which optimizes the path planning and navigation of the robot in unstructured environment. 

\section{Conclusion and Future Work}
In this work, we have presented off-road semantic segmentation (OFFSEG) framework  for fine semantic segmentation on two off-road datasets. OFFSEG shows promise for achieving good mIoU. The sub-class segmentation within the traversable region from OFFSEG can be used for robust scene understanding and optimized path planning for navigation through off-road environments. 
This framework can be extended to include other sub-classes which are not included in RELLIS-3D and RUGD within traversable region. Another interesting direction is to study the robustness of the approach under different climatic conditions changes as the vegetation and texture of an  off-road scene changes significantly compared to urban environments. 
We are in the process of generating a dataset from IISERB campus under diverse weather conditions.

\bibliographystyle{IEEEtran}
\bibliography{Reference}
\end{document}